\documentclass{article} 
\usepackage{tikz, pgf}
\usetikzlibrary{bayesnet}
\usepackage{multicol}
\usepackage{amsmath, amsthm, bm, amssymb}
\usepackage{nips12submit_e,times}
\usepackage{algpseudocode, algorithm}
\usepackage{caption,subcaption}
\usepackage[square, comma, sort&compress, numbers]{natbib}
\usepackage{wrapfig}
\DeclareMathOperator*{\argmin}{arg\,min}

\definecolor{DarkRed}{rgb}{0.75,0,0}
\definecolor{DarkGreen}{rgb}{0,0.5,0}
\definecolor{DarkBlue}{rgb}{0,0,0.5}
\definecolor{DarkPurple}{rgb}{0.5,0,0.5}

\newtheorem{theorem}{Theorem}
\newcommand{\dtk}{ {{\Delta t}_k} }
\title{Dynamic Clustering via Asymptotics of the Dependent Dirichlet Process
Mixture}

\author{
Trevor Campbell \\
MIT\\
Cambridge, MA 02139 \\
\texttt{\small tdjc@mit.edu} \\ \rule{.48\textwidth}{0pt}
\And
Miao Liu \\
Duke University\\
Durham, NC 27708 \\
\texttt{\small miao.liu@duke.edu} \\ \rule{.48\textwidth}{0pt}
\AND
Brian Kulis \\
Ohio State University \\
Columbus, OH 43210\\
\texttt{\small kulis@cse.ohio-state.edu} \\\rule{.32\linewidth}{0pt}
\And
Jonathan P.\ How \\
MIT \\
Cambridge, MA 02139 \\
\texttt{\small jhow@mit.edu} \\\rule{.32\linewidth}{0pt}
\And
Lawrence Carin\\
Duke University\\
Durham, NC 27708\\
\texttt{\small lcarin@duke.edu} \\\rule{.32\linewidth}{0pt}
}

\nipsfinalcopy 

\begin{document}

\maketitle

\begin{abstract}
This paper presents a novel algorithm,
based upon the dependent Dirichlet process mixture model (DDPMM), for clustering 
batch-sequential data containing an unknown number of evolving clusters. 
The algorithm is derived via a low-variance asymptotic analysis of the Gibbs sampling algorithm
for the DDPMM, and provides a hard clustering with convergence guarantees
similar to those of the k-means algorithm. Empirical results from a synthetic test with moving
Gaussian clusters and a test with real ADS-B aircraft trajectory data demonstrate that
the algorithm requires orders of magnitude less computational time than 
contemporary probabilistic and hard clustering algorithms, while providing
higher accuracy on the examined datasets.
\end{abstract}
\section{Introduction}
The Dirichlet process mixture model (DPMM) is a powerful tool for 
clustering data that enables the inference of an unbounded 
number of mixture components, and has been widely studied in the machine 
learning and statistics communities~\cite{Teh10_EML, Neal00_JCG, Blei06_BA,
Carvalho10_BA}. Despite its flexibility, 
it assumes the observations are exchangeable, and therefore that the 
data points have no inherent ordering that influences their labeling. This 
assumption is invalid for modeling temporally/spatially evolving phenomena, 
in which the order 
of the data points plays a principal role in creating meaningful clusters. 
The dependent Dirichlet process (DDP), originally formulated by 
MacEachern~\cite{MacEachern99_ASA}, provides a prior over such evolving 
mixture models, 
and is a promising tool for incrementally monitoring the dynamic 
evolution of the cluster structure within a dataset. 
More recently, a construction of the DDP built upon completely random
measures~\cite{Lin10_NIPS} led to the development of the dependent Dirichlet
process Mixture model (DDPMM) and a corresponding approximate
posterior inference Gibbs sampling algorithm. This model generalizes the DPMM by including 
birth, death and transition processes for the clusters in the model.

The DDPMM is a Bayesian nonparametric (BNP) model, part of an ever-growing 
class of probabilistic models
for which inference captures uncertainty in both the number of
parameters and their values. While these models are powerful in their capability
to capture complex structures in data without requiring explicit model selection, 
they suffer some practical shortcomings. Inference techniques for BNPs typically
fall into two classes: sampling methods (e.g., Gibbs sampling
\cite{Neal00_JCG} or particle learning~\cite{Carvalho10_BA}) and
optimization methods (e.g., variational
inference~\cite{Blei06_BA} or stochastic variational
inference~\cite{Hoffman12_AX}). Current methods based on sampling do not scale
well with the size of the dataset~\cite{DoshiVelez09_ICML}. Most optimization
methods require analytic derivatives and the selection of an upper bound on 
the number of clusters a priori, where the computational complexity increases
with that upper bound~\cite{Blei06_BA, Hoffman12_AX}. 
State-of-the-art techniques in both classes are not ideal for use in
contexts where performing inference quickly and reliably on large volumes of
streaming data  is crucial for timely decision-making, such as 
 autonomous robotic systems~\cite{Endres05_RSS, Luber04_RSS, Wang08_RSS}. On the other hand, many classical clustering methods
\cite{Lloyd82_IEEETIT, Pelleg00_ICML, Tibshirani01_JRSTATS} scale well with the size of the dataset
and are easy to implement, and advances have recently been made to 
capture the flexibility of Bayesian nonparametrics in such approaches
\cite{Kulis12_ICML}. However, as of yet there is no classical algorithm
that captures dynamic cluster structure with the same representational 
power as the DDP mixture model.  

This paper discusses the Dynamic Means algorithm, a novel hard clustering 
algorithm for spatio-temporal data derived from the low-variance asymptotic limit
of the Gibbs sampling algorithm for the dependent Dirichlet process Gaussian
mixture model. This algorithm captures the scalability and ease of
implementation of classical clustering methods, along with the representational
power of the DDP prior, and is guaranteed to converge to a local minimum of
a k-means-like cost function. The algorithm is significantly more
computationally tractable than Gibbs sampling, particle learning, and
variational inference for the DDP mixture model in practice, while providing
equivalent or better clustering accuracy on the examples presented. 
The performance and characteristics of the algorithm are demonstrated in
a test on synthetic data, with a comparison to those of
Gibbs sampling, particle learning and variational inference. Finally,
the applicability of the algorithm to real data is presented through
an example of clustering a spatio-temporal dataset of aircraft trajectories 
recorded across the United States.

\section{Background}
The Dirichlet process (DP) is a prior over mixture models, where the number of
mixture components is not known a priori\cite{Ferguson73_ANSTATS}. In general,
we denote $D \sim
\text{DP}(\mu)$, where $\alpha_\mu \in \mathbb{R}_+$ and $\mu : \Omega \rightarrow
\mathbb{R}_+, \int_\Omega \mathrm{d}\mu = \alpha_\mu$ are the 
\emph{concentration parameter} and \emph{base measure} of the DP, respectively.
If $D \sim \mathrm{DP}$,
then $D = \{(\theta_k, \pi_k)\}_{k=0}^\infty \subset \Omega \times
\mathbb{R}_+$, where $\theta_k \in \Omega$ and $\pi_k \in
\mathbb{R}_+$\cite{Sethuraman94_SS}. The
reader is directed to \cite{Teh10_EML} for a more thorough coverage of Dirichlet
processes.

The dependent Dirichlet process (DDP)\cite{MacEachern99_ASA}, an extension to the DP, is a prior over 
evolving mixture models. Given a Poisson process
construction\cite{Lin10_NIPS}, the DDP essentially forms a Markov chain of DPs
($D_1, D_2, \dots$), where
the transitions are governed by a set of three stochastic operations: 
Points $\theta_k$ may be added, removed, and may move during 
each step of the Markov chain. Thus, they become parameterized by time, denoted
by $\theta_{kt}$. In slightly more detail, if $D_{t}$ is the DP at time
step $t$, then the following procedure defines the generative model of $D_t$ conditioned
on $D_{t-1} \sim \mathrm{DP}(\mu_{t-1})$:
\begin{enumerate}
  \item{\textbf{Subsampling}:
    Define a function $q : \Omega \rightarrow [0,1]$. Then for each point
    $(\theta, \pi) \in D_{t-1}$, sample a Bernoulli distribution $b_\theta \sim
    \mathrm{Be}(q(\theta))$. Set $D'_t$ to be the collection of points $(\theta, \pi)$ such
    that $b_\theta = 1$, and renormalize the weights. Then $D'_t \sim
    \mathrm{DP}(q\mu_{t-1})$, where $(q\mu)(A) = \int_A q(\theta)\mu(d\theta)$.
  }
  \item{\textbf{Transition}: Define a distribution $T : \Omega \times \Omega
    \rightarrow \mathbb{R}_+$. For each point $(\theta, \pi)\in D'_t$, 
    sample $\theta' \sim T(\theta' | \theta)$, and set $D''_t$ to be 
    the collection of points $(\theta', \pi)$. Then $D''_t \sim
    \mathrm{DP}(Tq\mu_{t-1})$, where $(T\mu)(A) = \int_A\int_\Omega T(\theta' |
    \theta) \mu(d\theta)$.
  }
  \item{\textbf{Superposition}: Sample $F \sim \mathrm{DP}(\nu)$, and
    sample $(c_D, c_F) \sim \mathrm{Dir}(Tq\mu_{t-1}(\Omega), \nu(\Omega))$. 
    Then set $D_t$ to be the union of $(\theta, c_D\pi)$ for all $(\theta, \pi)
    \in D''_t$ and $(\theta, c_F\pi)$ for all $(\theta, \pi)\in F$. Thus,
    $D_t$ is a random convex combination of $D''_t$ and $F$, where
    $D_t \sim \mathrm{DP}(Tq\mu_{t-1} + \nu)$.
  }
\end{enumerate}
If the DDP is used as a prior over a mixture model, these three operations allow
new mixture components to arise over time, 
and old mixture components to exhibit dynamics
and perhaps disappear over time. As this is covered thoroughly in
\cite{Lin10_NIPS},
the mathematics of the underlying Poisson point process construction are 
not discussed in more depth in this work.
However, an important result of using such a construction is the development of
an explicit posterior for $D_t$ given observations of the points $\theta_{kt}$ at
timestep $t$. For each point $k$
that was observed in $D_{\tau}$ for some $\tau : 1\leq \tau \leq t$, define:
$n_{kt}\in\mathbb{N}$ as the number of observations of point $k$ in timestep
$t$; $c_{kt}\in\mathbb{N}$ as the number of past observations of point $k$ 
prior to timestep $t$,
i.e.~$c_{kt} = \sum_{\tau = 1}^{t-1} n_{k \tau}$; $q_{kt}\in(0, 1)$ as
the subsampling weight on point $k$ at timestep $t$; and $\dtk$ as the 
number of time steps that have elapsed since point $k$ was last observed. 
Further, let $\nu_t$ be 
the measure for unobserved points at time step $t$.
Then,
\begin{align}
  D_t | D_{t-1} &\sim \text{DP}\left(
  \nu_t +
  \sum_{k : n_{kt} = 0}q_{k t} c_{k t} T(\cdotp|\theta_{k(t-\dtk)}) + \sum_{k : n_{kt} >
0} (c_{k t} + n_{k t})\delta_{\theta_{kt}}\right)\label{eq:ddpposterior}
\end{align}
where $c_{kt} = 0$ for any point $k$ that was first observed during
timestep $t$. This posterior leads directly
to the development of a Gibbs sampling algorithm for the DDP,
whose low-variance asymptotics are discussed further below.

\section{Asymptotic Analysis of the DDP Mixture}
The dependent Dirichlet process Gaussian mixture model
(DDP-GMM)
serves as the foundation upon which the present work is built.
The generative model of a DDP-GMM at time step $t$ is
\begin{align}
  \begin{aligned}
    \{\theta_{kt}, \pi_{kt}\}_{k=1}^{\infty} &\sim \mathrm{DP}(\mu_t)\\
    \{z_{it}\}_{i=1}^{N_t} &\sim \mathrm{Categorical}(\{\pi_{kt}\}_{k=1}^\infty)\\
    \{y_{it}\}_{i=1}^{N_t} &\sim \mathcal{N}(\theta_{z_{it}t}, \sigma I)
  \end{aligned}
\end{align}
where $\theta_{kt}$ is the mean of cluster $k$, $\pi_{kt}$ 
is the categorical weight
for class $k$, $y_{it}$ is a $d$-dimensional observation vector, $z_{it}$ is a
cluster label for observation $i$, and $\mu_t$ is the base measure from
equation (\ref{eq:ddpposterior}). Throughout the rest of this paper, the subscript
$kt$ refers to quantities
related to cluster $k$ at time step $t$,
and subscript $it$ refers to quantities related to observation $i$ at time step
$t$.

The Gibbs sampling algorithm for the DDP-GMM iterates between sampling 
labels $z_{it}$ for datapoints $y_{it}$ given the set of parameters
$\{\theta_{kt}\}$, and sampling parameters $\theta_{kt}$ given each group of 
data $\{y_{it} : z_{it} = k\}$. Assuming the transition model $T$ is Gaussian,
and the subsampling function $q$ is constant, the functions and distributions
used in the Gibbs sampling algorithm are: 
the prior over cluster parameters, $\theta \sim \mathcal{N}(\phi, \rho I)$;
  the likelihood of an observation given its cluster parameter, $y_{it}
    \sim \mathcal{N}(\theta_{kt}, \sigma I)$; 
  the distribution over the transitioned cluster parameter given its 
    last known location after $\dtk$ time steps, 
    $\theta_{kt} \sim \mathcal{N}(\theta_{k(t-\dtk)}, \xi\dtk I)$;
  and the subsampling function $q(\theta) = q\in (0, 1)$.
Given these functions and distributions, the low-variance asymptotic 
limits (i.e.~$\sigma\rightarrow 0$) of these two steps are discussed in
the following sections.

\subsection{Setting Labels Given Parameters}
In the label sampling step, a datapoint $y_{it}$ can either create a new
cluster, join a current cluster, or revive an old, transitioned cluster. Using
the distributions defined previously, the label assignment probabilities are
\begin{align}
  p(z_{it} = k | \dots) &\propto \left\{\begin{array}{c c}
\alpha_t(2 \pi (\sigma+\rho))^{-d/2} \exp{\left({-\frac{|| y_{it}
  - \phi
||^2}{2 (\sigma+\rho)}}\right)} & k = K+1 \\
   (c_{kt} + n_{kt})(2 \pi \sigma)^{-d/2} \exp{\left({-\frac{||
   y_{it} - \theta_{kt}
 ||^2}{2 \sigma}}\right)} & n_{kt} > 0\\
 q_{kt} c_{kt} (2 \pi (\sigma+\xi\dtk))^{-d/2} \exp{\left({-\frac{|| y_{it} -
  \theta_{k(t-\dtk)}
||^2}{2 (\sigma+\xi\dtk)}}\right)} & n_{kt} = 0
  \end{array}\right.
\end{align}
where $q_{kt} = q^\dtk$ due to the fact that $q(\theta)$ is constant over
$\Omega$, and $\alpha_t = \alpha_\nu
\frac{1-q^t}{1-q}$ where $\alpha_\nu$ is the concentration parameter for 
the innovation process, $F_t$. The low-variance asymptotic limit
of this label assignment step
 yields meaningful assignments as long as $\alpha_\nu$, $\xi$, and
$q$ vary appropriately with $\sigma$; thus, setting $\alpha_\nu$, $\xi$, and $q$ as follows
(where $\lambda$, $\tau$ and $Q$ are positive constants):
\begin{align}
  \begin{array}{c c c}
    \alpha_\nu = (1+\rho/\sigma)^{d/2}\exp\left({-\frac{\lambda}{2\sigma}}\right),
&
  \xi = \tau \sigma,
&
  q =  \exp\left({-\frac{Q}{2 \sigma}}\right)
  \end{array}
\end{align}
 yields the following
assignments in the limit as $\sigma\rightarrow 0$:
\begin{align}
  z_{it} = \argmin_{k} \left\{ J_k \right\}, \, J_k =  
          \left\{\begin{array}{l l}
          ||y_{it} - \theta_{kt}||^2 \quad &\text{if $\theta_k$ instantiated}\\
          Q\dtk + \frac{|| y_{it} - \theta_{k(t-\dtk)}||^2}{\tau\dtk+1} \quad &\text{if
            $\theta_k$ old, uninstantiated}\\
          \lambda \quad &\text{if $\theta_k$ new} 
        \end{array}\right. .\label{eq:labelupdate}
\end{align}
  In this assignment step, $Q\dtk$ acts as a cost 
  penalty for reviving old clusters that increases with the time
  since the cluster was last seen, $\tau\dtk$ acts as a cost reduction to account for the possible motion of
clusters since they were last instantiated, and $\lambda$ acts as a cost penalty
for introducing a new cluster.

\subsection{Setting Parameters given Labels} 
In the parameter sampling step, the parameters are sampled using the
distribution 
\begin{align}
  \mathrm{p}(\theta_{kt} | \{y_{it} : z_{it} = k\}) \propto p(\{y_{it} : z_{it}
= k\} | \theta_{kt}) \mathrm{p}(\theta_{kt})
\end{align}
There are two cases to consider when setting a
parameter $\theta_{kt}$. Either $\dtk = 0$ and the cluster is new in the current time step, or
$\dtk > 0$ and the cluster was previously 
created, disappeared for some amount of time, and then was revived in the
current time step. 
\paragraph*{New Cluster}
Suppose cluster $k$ is being newly created. In this case,
$\theta_{kt} \sim \mathcal{N}(\phi, \rho)$.
Using the fact that a normal
prior is conjugate a normal likelihood, the closed-form posterior for
$\theta_{kt}$ is
\begin{align}
\begin{aligned}
  \theta_{kt} | \{y_{it}: z_{it} = k\} &\sim
  \mathcal{N}\left( \theta_{\text{post}},
  \sigma_{\text{post}}\right)\\ 
  \theta_{\text{post}} =  \sigma_{\text{post}}\left(\frac{\phi}{\rho} + 
  \frac{\sum_{i=1}^{n_{kt}}
y_{it}}{\sigma}\right)&, \,
\sigma_{\text{post}} =
\left(\frac{1}{\rho} + \frac{n_{kt}}{\sigma}\right)^{-1}
\end{aligned}
\end{align}
 Then letting $\sigma\rightarrow
0$, 
\begin{align}
  \theta_{kt} &= \frac{\left(\sum_{i=1}^{n_{kt}} y_{it}\right)}{n_{kt}}
  \overset{\tt def}{=} m_{kt}
\end{align}
where $m_{kt}$ is the mean of the observations in the current timestep.

\paragraph*{Revived Cluster} Suppose there are $\dtk$ 
time steps where cluster $k$ was not observed, but
there are now $n_{kt}$ data points with mean $m_{kt}$ assigned 
to it in this time step. In this case,
\begin{align}
\mathrm{p}(\theta_{kt}) = \int_{\theta} T(\theta_{kt}
|\theta)\mathrm{p}(\theta)\,
\mathrm{d}\theta, \, \, \theta \sim \mathcal{N}(\theta', \sigma').
\end{align}
Again
using conjugacy of normal likelihoods and priors,
\begin{align}
\begin{aligned}
  \theta_{kt} | \{y_{it}: z_{it} = k\} &\sim
  \mathcal{N}\left( \theta_{\text{post}},
  \sigma_{\text{post}}\right)\\ 
  \theta_{\text{post}} =  \sigma_{\text{post}}\left(\frac{\theta'}{\xi \dtk +
  \sigma'} + 
  \frac{\sum_{i=1}^{n_{kt}}
y_{it}}{\sigma}\right)&, \,
\sigma_{\text{post}} =
\left(\frac{1}{\xi \dtk + \sigma'} + \frac{n_{kt}}{\sigma}\right)^{-1}
\end{aligned}\label{eq:transpost}
\end{align}
Similarly to the label assignment step, let $\xi = \tau \sigma$. Then as long as
$\sigma' = \sigma/w$, $w > 0$ 
(which holds if equation (\ref{eq:transpost}) is used
to recursively keep track of the parameter posterior), taking the asymptotic limit 
of this as $\sigma\rightarrow 0$ yields:
\begin{align}
  \theta_{kt} &= \frac{\theta' (w^{-1}+\dtk \tau)^{-1} +
  n_{kt}
  m_{kt} } { (w^{-1}+\dtk \tau)^{-1} +  n_{kt}}
\end{align}
that is to say, the revived $\theta_{kt}$ is a weighted 
average of estimates using current
timestep data and previous timestep data. $\tau$ controls how much the current 
data is favored - as $\tau$ increases, the weight on current data increases,
which is explained by the fact
that our uncertainty in where the old $\theta'$ transitioned to increases with
$\tau$. It is also noted that if $\tau = 0$, this reduces to a simple weighted
average using the amount of data collected as weights.

\paragraph{Combined Update} Combining the updates for new cluster parameters and
old transitioned cluster parameters yields a recursive update scheme:
\begin{align}
  \begin{aligned}
    \theta_{k0} &= m_{k0}\\
    w_{k0} &= n_{k0}
  \end{aligned}\quad \text{and} \quad
  \begin{aligned}
    \gamma_{kt} &= \left( (w_{k(t-\dtk)})^{-1} + \dtk \tau\right)^{-1}\\
    \theta_{kt} &= \frac{\theta_{k(t-\dtk)}\gamma_{kt} + n_{kt}
    m_{kt}} {\gamma_{kt} + n_{kt}}\\
    w_{kt} &= \gamma_{kt} + n_{kt}
\end{aligned} \label{eq:paramupdate}
\end{align}
where time step $0$ here corresponds to when the cluster is first created.
An interesting interpretation of this update is that it behaves like 
a standard Kalman filter, in which $w_{kt}^{-1}$ serves as the 
current estimate variance, $\tau$ serves as the 
process noise variance, and $n_{kt}$
serves as the inverse of the measurement variance.

\section{The Dynamic Means Algorithm}
In this section, some further notation is required for brevity:
\begin{align}
\begin{aligned}
\mathcal{Y}_t = \{y_{it}\}_{i=1}^{N_t}&, \quad
\mathcal{Z}_t = \{z_{it}\}_{i=1}^{N_t}\\
\mathcal{K}_t = \{(\theta_{kt}, w_{kt}) : n_{kt}>0\}&, \quad
\mathcal{C}_t = \{(\dtk, \theta_{k(t-\dtk)}, w_{k(t-\dtk)})\}
\end{aligned}
\end{align}
where $\mathcal{Y}_t$ and $\mathcal{Z}_t$ are the sets of observations and
labels at time step $t$, $\mathcal{K}_t$ is the set of 
currently active clusters (some are new with $\dtk = 0$, and some are
revived with $\dtk > 0$), and $\mathcal{C}_t$ is the set
of old cluster information.

\subsection{Algorithm Description}
As shown in the previous section, the low-variance asymptotic limit of the
DDP Gibbs sampling algorithm is a deterministic observation label 
update (\ref{eq:labelupdate}) followed by a 
deterministic, weighted least-squares parameter update
(\ref{eq:paramupdate}). Inspired by the original K-Means algorithm, applying
these two updates iteratively yields an algorithm which
clusters a set of observations at
a single time step given cluster means and weights from past time steps
(Algorithm \ref{alg:dynmean_inner}).
Applying Algorithm \ref{alg:dynmean_inner} to a sequence of batches of data 
yields a clustering procedure that is able to track a set of
dynamically evolving clusters (Algorithm \ref{alg:dynmean}), 
and allows new clusters to emerge and old clusters to be forgotten. 
While this is the primary application of Algorithm \ref{alg:dynmean}, the 
sequence of batches need not be a temporal sequence. For example, Algorithm
\ref{alg:dynmean} may be used as an any-time clustering algorithm for
large datasets, where the sequence of batches is generated by selecting
random subsets of the full dataset. 

The \textsc{AssignParams} function is exactly the
update from equation (\ref{eq:paramupdate}) applied to 
each $k\in\mathcal{K}_t$. Similarly, the \textsc{AssignLabels} function applies
the update from equation (\ref{eq:labelupdate}) to each observation; 
however, in the case that a new cluster is created or an old one is 
revived by an observation, 
\textsc{AssignLabels} also creates a parameter for that new cluster
based on the parameter update equation (\ref{eq:paramupdate}) with 
that single observation. Note that the performance of the algorithm
depends on the order in which \textsc{AssignLabels} assigns labels. 
Multiple random restarts of the algorithm with different assignment orders 
may be used to mitigate this dependence.
The \textsc{UpdateC} function 
is run after clustering observations from each time step, and
constructs $\mathcal{C}_{t+1}$ by setting $\dtk=1$ for any new
or revived cluster, and by incrementing $\dtk$ for any
old cluster that was not revived:
\begin{align}
  \begin{aligned}
  \mathcal{C}_{t+1} &= \{(\dtk+1, \theta_{k(t-\dtk)}, w_{k(t-\dtk)}) : k\in
  \mathcal{C}_t, k\notin \mathcal{K}_t\} \cup \{(1, \theta_{kt},
  w_{kt}) : k\in \mathcal{K}_t\}
\end{aligned}
\end{align}

An important question is whether this algorithm is guaranteed to 
converge while clustering data in each time step. Indeed, it is;
Theorem \ref{thm:monodecrease} shows that a particular cost function $L_t$
monotonically decreases under the label and parameter updates
(\ref{eq:labelupdate}) and (\ref{eq:paramupdate}) at each time step. 
Since $L_t \geq 0$, and it is monotonically decreased by Algorithm
$\ref{alg:dynmean_inner}$, the algorithm converges. Note that the Dynamic Means
is only guaranteed to converge to a local optimum, similarly to the
k-means\cite{Lloyd82_IEEETIT} and
DP-Means\cite{Kulis12_ICML} algorithms.

\begin{table}[t!]
\vspace*{-.1in}
\hspace*{.2in}
\begin{minipage}[t]{.4\textwidth}
  \begin{algorithm}[H]
      \captionsetup{font=scriptsize}
      \caption{Dynamic Means}\label{alg:dynmean}
      \scriptsize
  \begin{algorithmic}
    \Require $\left\{\mathcal{Y}_t\right\}_{t=1}^{t_{f}}$, 
            $Q$, $\lambda$, $\tau$
    \State $\mathcal{C}_1 \gets \emptyset$
    \For{$t = 1 \to t_f$}
    \State $(\mathcal{K}_t, \mathcal{Z}_t,
    L_t)\gets$\Call{Cluster}{$\mathcal{Y}_t$,
    $\mathcal{C}_t$, 
    $Q$, $\lambda$, $\tau$}
    \State $\mathcal{C}_{t+1}\gets$\Call{UpdateC}{$\mathcal{Z}_t$, $\mathcal{K}_t$,
    $\mathcal{C}_t$}
    \EndFor
    \State \Return $\{\mathcal{K}_t, \mathcal{Z}_t, L_t\}_{t=1}^{t_f}$
  \end{algorithmic}
\end{algorithm}
\end{minipage}
\hspace{0.1in}
\begin{minipage}[t]{.4\textwidth}
  \begin{algorithm}[H]
    \captionsetup{font=scriptsize}
    \caption{\textsc{Cluster}}\label{alg:dynmean_inner}
      \scriptsize
  \begin{algorithmic}
    \Require $\mathcal{Y}_t$, $\mathcal{C}_t$, $Q$, $\lambda$, $\tau$
    \State $\mathcal{K}_t \gets \emptyset$, 
          $\mathcal{Z}_t \gets \emptyset$, 
          $L_0 \gets \infty$
    \For{$n = 1\to\infty$}
    \State $(\mathcal{Z}_t,
    \mathcal{K}_t)\gets$\Call{AssignLabels}{$\mathcal{Y}_t$, $\mathcal{Z}_t$,
    $\mathcal{K}_t$, $\mathcal{C}_t$}
    \State $(\mathcal{K}_t, L_n)\gets$\Call{AssignParams}{$\mathcal{Y}_t$,
    $\mathcal{Z}_t$, $\mathcal{C}_t$}
    \If{$L_n = L_{n-1}$}
      \State \Return $\mathcal{K}_t$, $\mathcal{Z}_t$, $L_n$
    \EndIf
    \EndFor
  \end{algorithmic}
  \end{algorithm}
\end{minipage}
\begin{minipage}[t]{.1\textwidth}\hspace*{\textwidth}\end{minipage}
\end{table}

\begin{theorem}\label{thm:monodecrease}
  Each iteration in Algorithm \ref{alg:dynmean_inner} 
  monotonically decreases the cost 
  function $L_t$, where
\begin{align}
  L_t &=\sum_{k\in \mathcal{K}_t}
  \left(\overbrace{\lambda\left[\dtk =
  0\right]}^{\text{New Cost}}+\overbrace{Q\dtk}^{\text{Revival Cost}}+
  \overbrace{\gamma_{kt}||\theta_{kt} -
  \theta_{k(t-\dtk)}||^2_2 +  \sum_{\begin{subarray}{c}y_{it}\in\mathcal{Y}_t \\
    z_{it} = k\end{subarray}} ||y_{it} -
\theta_{kt}||_2^2}^{\text{Weighted-Prior Sum-Squares Cost}}\right)\label{eqn:DDPobjective}
\end{align}
\end{theorem}
The cost function is comprised of a number of components for each
currently active cluster $k\in\mathcal{K}_t$: A penalty
for new clusters based on $\lambda$, a penalty for old clusters based
on $Q$ and $\dtk$, and finally a prior-weighted sum of squared
distance cost for all the 
observations in cluster $k$. It is noted that for new clusters, $\theta_{kt} =
\theta_{k(t-\dtk)}$ since $\dtk = 0$, so the least squares cost is
unweighted. The \textsc{AssignParams} function calculates this cost function 
in each iteration of Algorithm \ref{alg:dynmean_inner}, and the
algorithm terminates once the cost function does not decrease during
an iteration.

\subsection{Reparameterizing the Algorithm}
In order to use the Dynamic Means 
algorithm, there are three free parameters to
select: $\lambda$, $Q$, and $\tau$. While $\lambda$ represents
how far an observation can be from a cluster before it
is placed in a new cluster, and thus can be tuned 
intuitively, $Q$ and $\tau$ are not so straightforward. The 
parameter $Q$ represents a conceptual added distance from any data point
to a cluster for every time step that the cluster is not observed.
The parameter $\tau$ represents a conceptual reduction of distance from any
data point to a cluster for every time step that 
the cluster is not observed. How these two quantities affect
the algorithm, and how they interact with the setting of $\lambda$,
is hard to judge.

Instead of picking $Q$ and $\tau$ directly, the algorithm may be
reparameterized by picking $ N_Q, k_\tau \in \mathbb{R}_+$, $N_Q > 1$,
$k_\tau \geq 1$, and given
a choice of $\lambda$, setting
\begin{align}
\begin{aligned}
  Q =& \lambda/N_Q \quad
  \tau = \frac{N_Q(k_\tau-1) + 1}{N_Q-1}.
\end{aligned}
\end{align}
If $Q$ and $\tau$ are set in this manner, $N_Q$ represents the number
(possibly fractional) of
time steps a cluster can be unobserved before the label update (\ref{eq:labelupdate}) will never revive
that cluster, and $k_\tau \lambda$ 
represents the maximum squared distance away from a cluster 
center such that after a single time step, the
label update (\ref{eq:labelupdate}) will
revive that cluster. As $N_Q$ and $k_\tau$ 
are specified in terms of concrete algorithmic behavior, they are 
intuitively easier to set than $Q$ and $\tau$.

\section{Related Work}
Prior k-means clustering algorithms that determine the number of clusters present
in the data have primarily involved a method for
iteratively modifying k using various statistical criteria~\cite{Ishioka00_IDEAL, Pelleg00_ICML, Tibshirani01_JRSTATS}. In contrast, this
work derives this capability from a Bayesian
nonparametric model, similarly to the DP-Means algorithm~\cite{Kulis12_ICML}. In this sense, the relationship between the Dynamic 
Means algorithm and the dependent Dirichlet
process~\cite{Lin10_NIPS} is exactly that between the DP-Means
algorithm and Dirichlet process~\cite{Ferguson73_ANSTATS}, where the Dynamic
Means algorithm may be seen as an extension to the DP-Means that handles 
sequential data with time-varying cluster parameters. 
MONIC~\cite{Spiliopoulou06_KDD} and MC3~\cite{Kalnis05_SSTD} have the
capability to monitor time-varying clusters; however, these methods
require datapoints to be identifiable across timesteps, and determine cluster
similarity across timesteps via the commonalities between label assignments. The
Dynamic Means algorithm does not require such information, and tracks clusters
essentially based on similarity of the parameters across timesteps. Evolutionary
clustering~\citep{Chakraborti06_KDD,Xu12_DMKD}, similar to Dynamic Means, minimizes an objective consisting of a cost for clustering the 
present data set and a cost related to the comparison between the current clustering
and past clusterings. The present 
work can be seen as a theoretically-founded extension of this class of 
algorithm that provides methods for automatic and adaptive prior weight
selection, forming correspondences between old and
current clusters, and for deciding when to introduce new clusters. 
Finally, some sequential Monte-Carlo methods (e.g.~particle
learning~\cite{Carvalho10_SS} or multi-target
tracking~\cite{Hue02_TAES,Vermaak03_ICCV}) 
can be adapted for use in the present context, but suffer the drawbacks
typical of particle filtering methods. 

\section{Applications}
\subsection{Synthetic Gaussian Motion Data}
\begin{figure}
\captionsetup{font=scriptsize}
\begin{center}
  \begin{subfigure}[t]{.32\linewidth}
  \includegraphics[width=\textwidth]{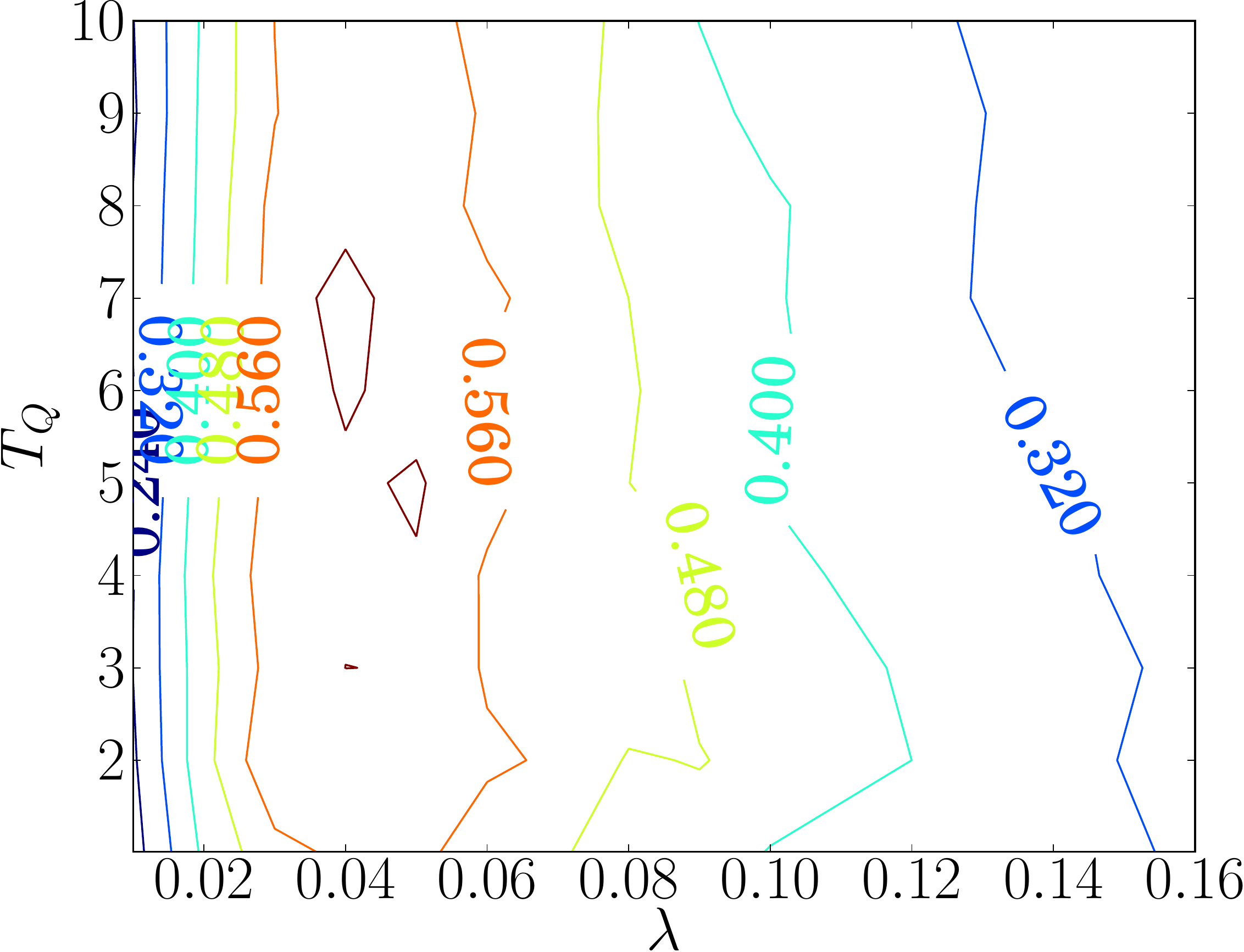}
    \caption{}\label{fig:TuneDynM-LamTQ}
  \end{subfigure}
  \begin{subfigure}[t]{.32\linewidth}
  \includegraphics[width=\textwidth]{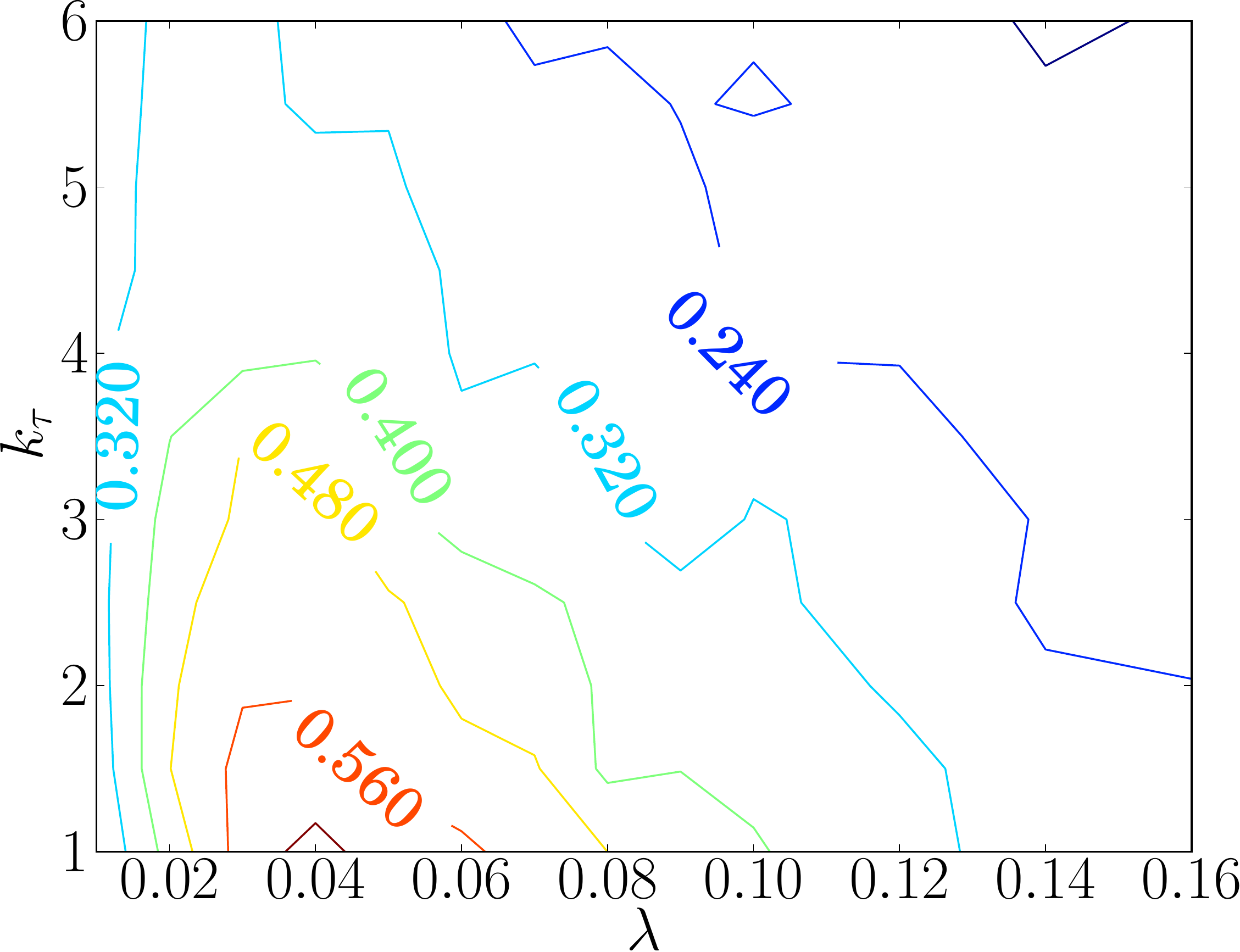}
    \caption{}\label{fig:TuneDynM-LamKT}
  \end{subfigure}
  \begin{subfigure}[t]{.295\linewidth}
  \includegraphics[width=\textwidth]{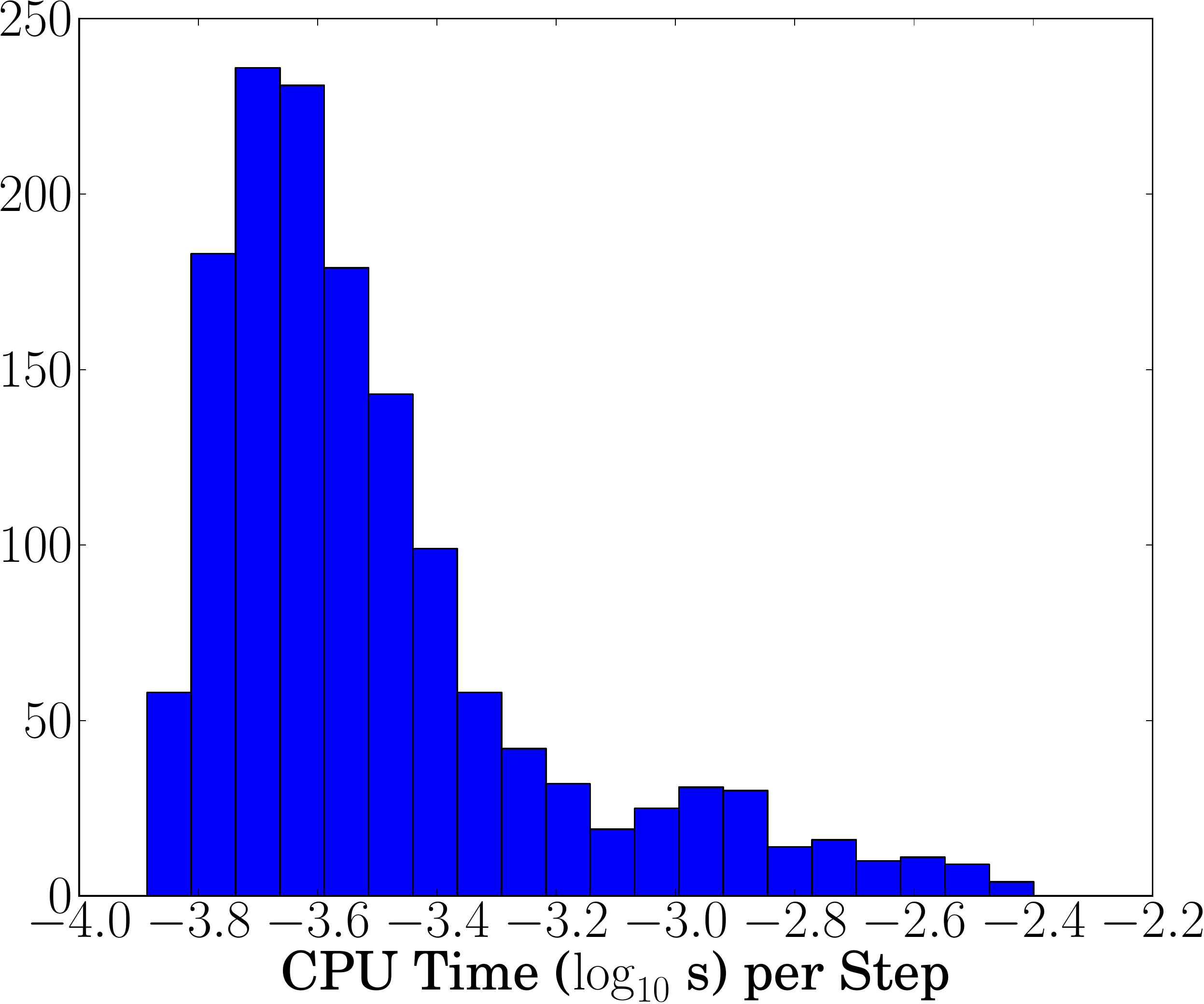}
    \caption{}\label{fig:TuneDynM-CpuT}
  \end{subfigure}
  \begin{subfigure}[t]{.34\linewidth}
  \includegraphics[trim=0 0 0 0, clip=true,width=\textwidth]{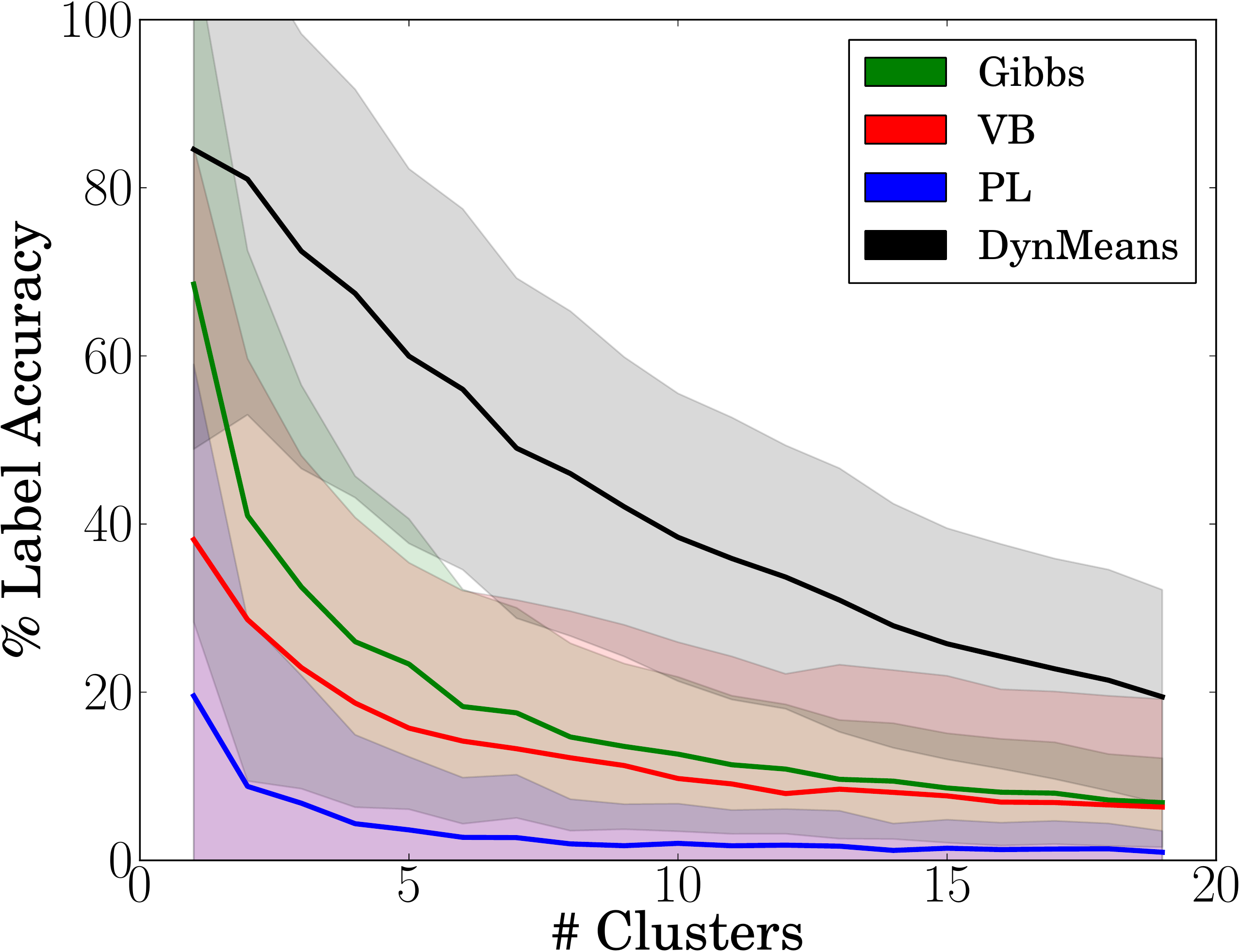}
    \caption{}\label{fig:DMvsDDP-acc}
  \end{subfigure}
  \begin{subfigure}[t]{.32\linewidth}
    \includegraphics[trim=0 0 0 0, clip=true,width=\textwidth]{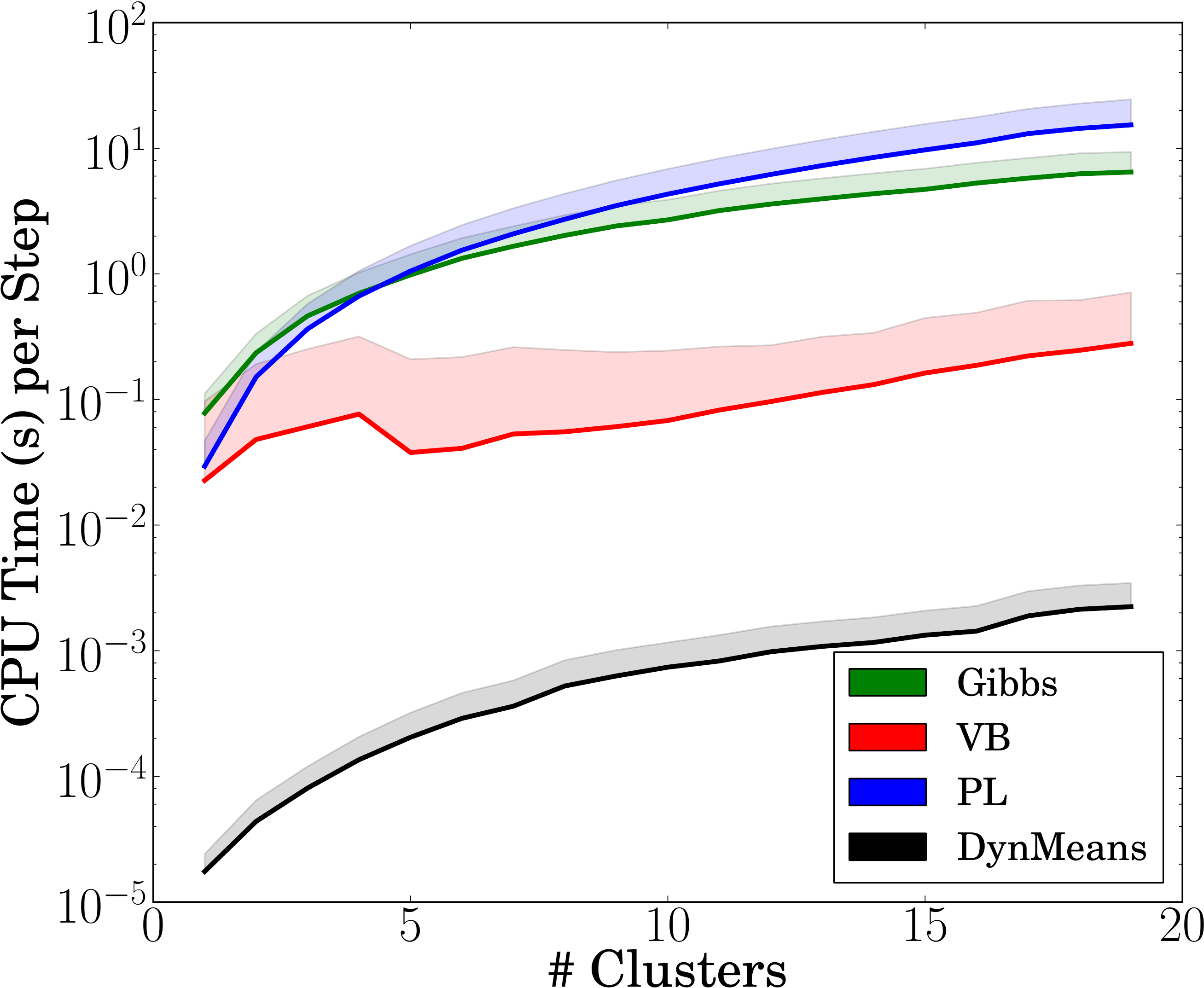}
    \caption{}\label{fig:DMvsDDP-time}
  \end{subfigure}
  \begin{subfigure}[t]{.32\linewidth}
    \includegraphics[trim=0 0 0 0,clip=true,width=\textwidth]{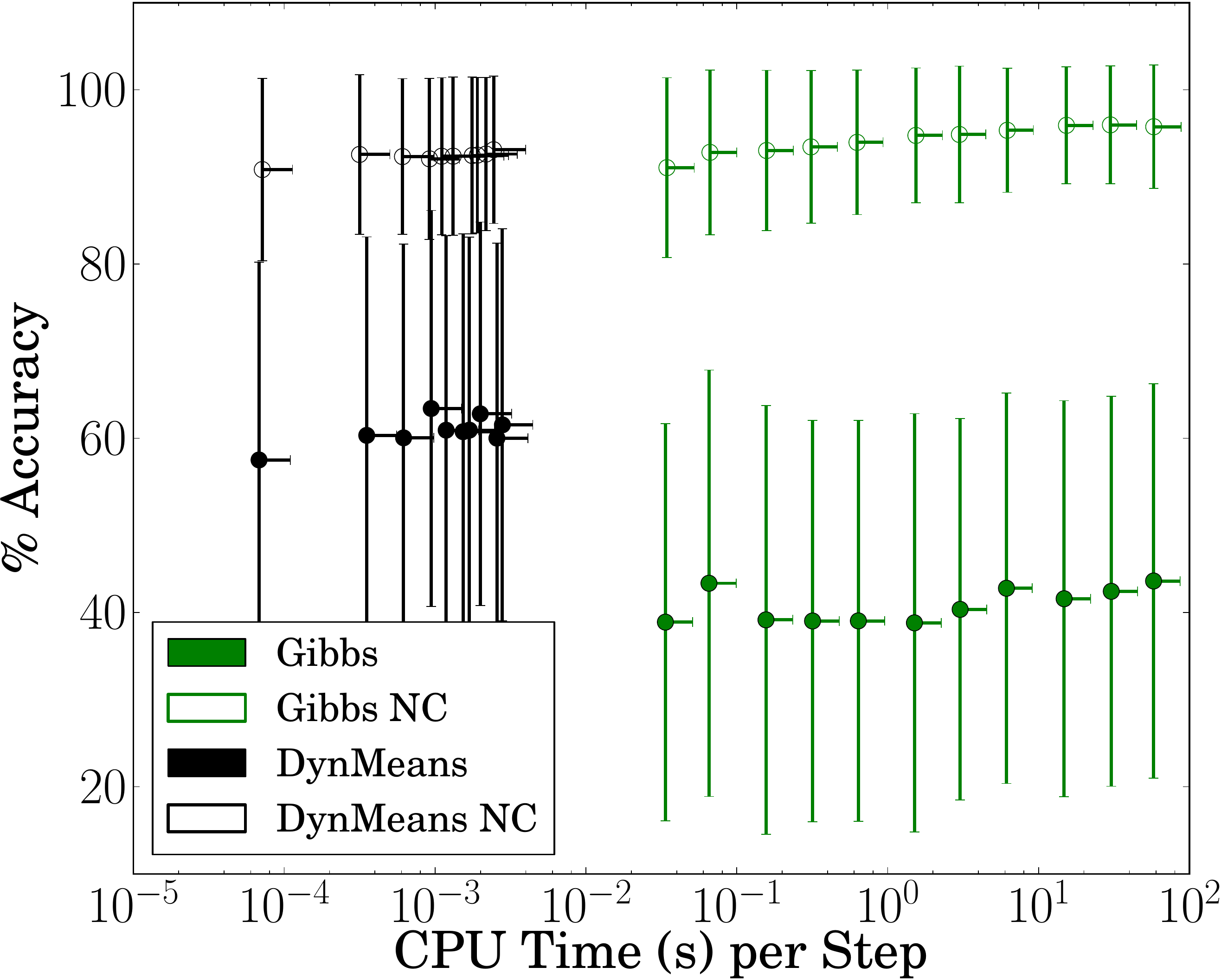}
    \caption{}\label{fig:DMvsDDP-accvtime}
  \end{subfigure}
  \caption{(\ref{fig:TuneDynM-LamTQ} - \ref{fig:TuneDynM-CpuT}): Accuracy
  contours and CPU time histogram for the Dynamic Means algorithm.
  (\ref{fig:DMvsDDP-acc} - \ref{fig:DMvsDDP-time}): Comparison
  with Gibbs sampling, variational inference, and particle learning. Shaded
  region indicates $1\sigma$ interval; in (\ref{fig:DMvsDDP-time}), only upper
  half is shown. (\ref{fig:DMvsDDP-accvtime}): Comparison of accuracy when
  enforcing (Gibbs, DynMeans) and not enforcing (Gibbs NC, DynMeans NC) correct
  cluster tracking.
}\label{fig:SynthTest}
\end{center}
\end{figure}

In this experiment, moving Gaussian clusters on $[0,1]\times[0,1]$ 
were generated synthetically
over a period of 100 time steps. 
In each step, there was some number
of clusters, each having 15 data points. The data points were sampled from
a symmetric Gaussian distribution with a standard deviation of 0.05. 
Between time steps, the cluster centers moved randomly, with displacements
sampled from the same distribution. At each time 
step, each cluster had a 0.05 probability of being destroyed.

This data was clustered with Dynamic Means (with 3 random assignment ordering restarts), 
DDP-GMM Gibbs sampling~\cite{Lin10_NIPS}, variational
inference~\cite{Blei06_BA}, and particle learning~\cite{Carvalho10_BA} 
on a computer with an Intel i7 processor and 16GB of memory. First, the number of clusters was
fixed to $5$, and the parameter space of each algorithm was searched for the
best possible cluster label accuracy (taking into 
account correct cluster tracking across time steps). 
The results of this parameter sweep for
the Dynamic Means algorithm with $50$ trials at each parameter setting are
shown in Figures \ref{fig:TuneDynM-LamTQ}--\ref{fig:TuneDynM-CpuT}. Figures~\ref{fig:TuneDynM-LamTQ} and \ref{fig:TuneDynM-LamKT} show how the average clustering
accuracy varies with the parameters after fixing either $k_{\tau}$ or $T_Q$ to their
values at the maximum accuracy parameter setting over the full space. The Dynamic Means
algorithm had a similar robustness with respect to variations in its parameters as the
comparison algorithms. The histogram in
Figure \ref{fig:TuneDynM-CpuT} demonstrates that the clustering speed is robust
to the setting of parameters. The speed of Dynamic Means, coupled with the
smoothness of its performance with respect to its parameters, makes it
well suited for automatic tuning~\cite{Snoek12_NIPS}.

Using the best parameter setting for each algorithm, 
the data as described above were clustered in 50 trials with a varying number of
clusters present in the data. For the Dynamic Means algorithm, 
parameter values $\lambda = 0.04$, $T_Q =
6.8$, and $k_\tau = 1.01$ were used, and the algorithm was again given 3
attempts with random labeling assignment orders, where the lowest cost solution
of the 3 was picked to proceed to the next time step. For the other algorithms, 
the parameter values $\alpha = 1$ and $q = 0.05$ were used,
with a Gaussian transition distribution variance of 0.05. The number of samples
for the Gibbs sampling algorithm was 5000 with one recorded for every 5 samples, the number of particles for the particle learning algorithm was 100, and the variational
inference algorithm was run to a tolerance of $10^{-20}$ with the maximum number of
iterations set to 5000. 

In Figures \ref{fig:DMvsDDP-acc} and
\ref{fig:DMvsDDP-time}, the 
labeling accuracy and clustering time (respectively) 
for the algorithms is shown. The sampling algorithms were
handicapped to generate 
Figure \ref{fig:DMvsDDP-acc}; the best posterior sample 
in terms of labeling accuracy 
was selected at each time step, which required knowledge of the true
labeling. Further, the accuracy computation included
enforcing consistency across timesteps, to allow tracking individual
cluster trajectories. If this is not enforced (i.e.~accuracy 
considers each time step independently), the other algorithms 
provide accuracies more comparable to those of the Dynamic
Means algorithm. This effect is demonstrated in 
Figure \ref{fig:DMvsDDP-accvtime}, which shows the time/accuracy tradeoff 
for Gibbs sampling (varying the number of samples) and Dynamic Means
(varying the number of restarts). These examples illustrate that Dynamic
Means outperforms standard inference algorithms in both label accuracy and
computation time for cluster tracking problems.


\subsection{Aircraft Trajectory Clustering}

\begin{figure}[t!]
\captionsetup{font=scriptsize}
\begin{center}
  \begin{subfigure}[c]{.28\textwidth}
  \includegraphics[width=\textwidth, trim=60 210 80 60,
  clip]{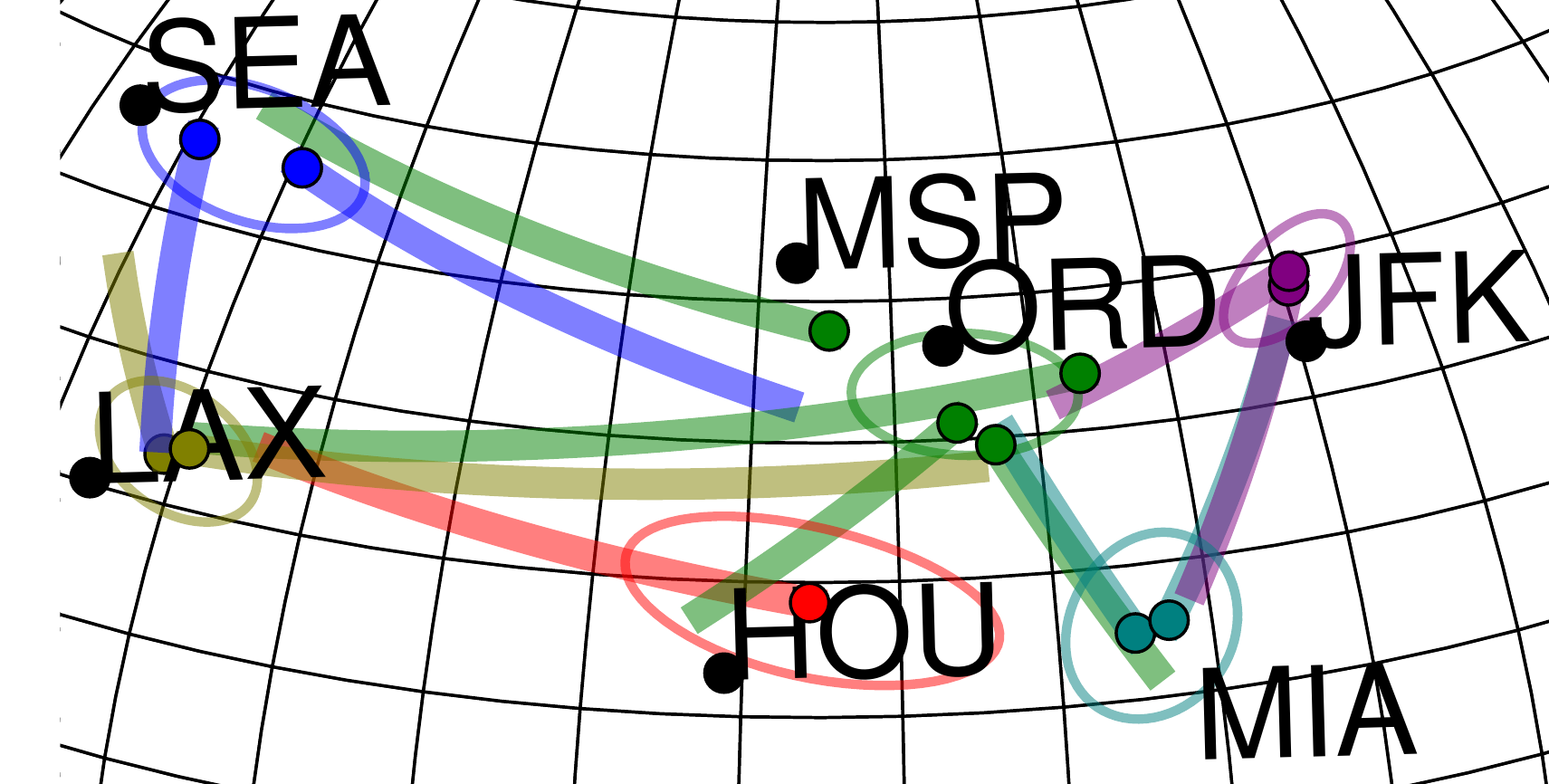}
\end{subfigure}
  \begin{subfigure}[c]{.7\textwidth}
    \includegraphics[width=\textwidth]{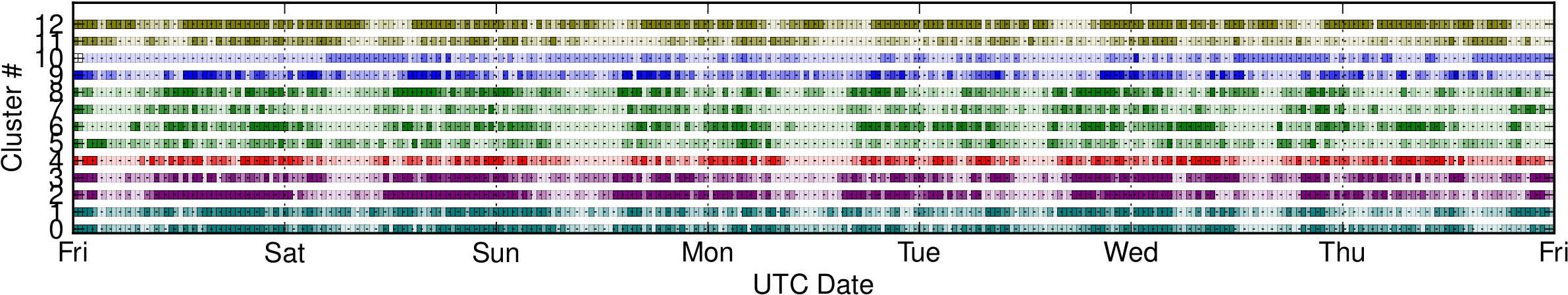}
  \end{subfigure}
  \caption{Results of the GP aircraft trajectory clustering. Left: A map
    (labeled with major US city airports) showing the overall aircraft 
    flows for 12 trajectories, with colors and 1$\sigma$ confidence ellipses corresponding to takeoff
    region (multiple clusters per takeoff region), colored dots indicating
    mean takeoff position for each cluster, and lines indicating the mean
    trajectory for each cluster.
    Right: A track of plane counts for the 12 clusters during the week, with color intensity proportional to
  the number of takeoffs at each time.}\label{fig:GPresults}
  \end{center}
\end{figure}



In this experiment, the Dynamic Means algorithm was used to find the
typical spatial and temporal patterns in the motions of commercial aircraft.
Automatic dependent surveillance-broadcast (ADS-B) data,
including plane identification, timestamp, latitude, longitude, heading and speed, was
collected from all transmitting planes across the United States
during the week from 2013-3-22 1:30:0 to 2013-3-28 12:0:0 UTC. 
Then,
individual ADS-B messages were connected together based on their plane
identification and timestamp to form trajectories, and erroneous trajectories
were filtered based on reasonable spatial/temporal bounds, 
yielding 17,895 unique trajectories.
Then, for each trajectory, a Gaussian process was trained using the latitude and
longitude of each ADS-B point along the trajectory as the inputs and the North 
and East components of plane velocity at those points as
the outputs. Next, the mean latitudinal and longitudinal velocities from the
Gaussian process were queried
for each point on a regular lattice across the USA (10 latitudes and 20
longitudes), and used to create a 400-dimensional feature vector for each
trajectory. Of the resulting 17,895 feature vectors, 600 were hand-labeled
(each label including a confidence weight in $[0, 1]$). The feature vectors 
were clustered using the DP-Means algorithm on the entire dataset in a single
batch, and using Dynamic Means / DDPGMM Gibbs sampling (with 50 samples) with half-hour takeoff
window batches.

\begin{wraptable}[7]{r}{.325\textwidth}
  \centering \vspace*{-.2in}
  \captionsetup{font=scriptsize}
  \caption{Mean computational time \& accuracy on hand-labeled aircraft
  trajectory data}\label{tab:GPresults}
  {\small
  \begin{tabular}{l|c|l}
    \textbf{Alg.} & \textbf{\% Acc.} & \textbf{Time (s)}\\
    \hline
\rule{0pt}{10pt}DynM & $55.9$ & $2.7\times 10^2$\\ 
    DPM &  $55.6$& $3.1\times 10^3$\\ 
    Gibbs & $36.9$ & $1.4\times 10^4$ 
  \end{tabular}
}
\end{wraptable}
The results of this exercise
are provided in Figure \ref{fig:GPresults} and Table \ref{tab:GPresults}. 
Figure \ref{fig:GPresults} shows the spatial and temporal properties 
of the 12 most popular clusters discovered by Dynamic Means, demonstrating
that the algorithm successfully identified major flows of commercial
aircraft across the US. Table \ref{tab:GPresults} corroborates
these qualitative results with a quantitative comparison of
the computation time and accuracy for the three algorithms tested over 20 trials.
The confidence-weighted accuracy was computed by taking the ratio between the sum of the weights for
correctly labeled points and the sum of all weights. The DDPGMM Gibbs sampling
algorithm was handicapped as described in the synthetic experiment section. Of
the three algorithms,
Dynamic Means provided the highest labeling accuracy, while requiring
orders of magnitude less computation time than both DP-Means and DDPGMM Gibbs
sampling.

%
%

\section{Conclusion}
This work developed a clustering algorithm for batch-sequential
data containing temporally evolving clusters, derived from a low-variance
asymptotic analysis of the Gibbs sampling algorithm for the dependent Dirichlet
process mixture model. Synthetic and real data experiments demonstrated that
the algorithm requires orders of magnitude less computational time than 
contemporary probabilistic and hard clustering algorithms, while providing
higher accuracy on the examined datasets. The speed of inference coupled with the
convergence guarantees provided yield an algorithm which is suitable for use in
time-critical applications, such as online model-based autonomous
planning systems.

\subsubsection*{Acknowledgments}
This work was supported by NSF award IIS-1217433 and ONR MURI grant
N000141110688.

\bibliographystyle{unsrt}
{\small
\bibliography{main}
}
\end{document}